\documentclass[twoside]{article}
\pdfoutput=1

\usepackage[accepted]{aistats2018}

\usepackage{color} 

\usepackage[T1]{fontenc}    
\usepackage{lmodern}

\usepackage{hyperref}       
\usepackage{url}            
\usepackage{booktabs}       
\usepackage{amsfonts}       
\usepackage{nicefrac}       
\usepackage{microtype}      
\usepackage{amsmath,amssymb,amsthm,mathrsfs,amsfonts,dsfont}
\usepackage{graphicx}
\usepackage{wrapfig}
\usepackage{xspace}
\usepackage{soul}  
\usepackage{nicefrac}
\usepackage[authoryear,round]{natbib}

\newcommand{\thetav}{\boldsymbol{\theta}}
\newcommand{\etav}{\boldsymbol{\eta}}
\newcommand{\xiv}{\boldsymbol{\xi}}

\newcommand{\yv}{\mathbf{y}}
\newcommand{\zv}{\mathbf{z}}
\newcommand{\mv}{\mathbf{m}}
\newcommand{\Sv}{\mathbf{S}}
\newcommand{\Tv}{\mathbf{t}}
\newcommand{\Fv}{\mathbf{F}}

\newcommand{\uv}{\mathbf{u}}
\newcommand{\inv}{^{-1}}
\newcommand{\T}{^{\top }}
\newcommand{\E}{\mathbb{E}}

\newcommand{\dd}[2]{\frac{\partial #1}{\partial #2}}

\newcommand{\KL}{\textsc{kl}}

\newcommand{\R}{\mathbb{R}}

\newcommand{\Pv}{\mathbf{P}}
\newcommand{\gv}{\mathbf{g}}
\newcommand{\deltav}{\boldsymbol{\delta}}

\newcommand{\Thetav}{\boldsymbol{\Theta}}

\DeclareRobustCommand{\LBFGS}{LBFGS\xspace}

\DeclareRobustCommand{\stepsize}{step size\xspace}
\DeclareRobustCommand{\linesearch}{line search\xspace}
\DeclareRobustCommand{\elementwise}{elementwise\xspace}
\DeclareRobustCommand{\supplementarymaterial}{supplementary material\xspace}
\DeclareRobustCommand{\studentT}{student-t\xspace}
\DeclareRobustCommand{\StudentT}{Student-t\xspace}

\DeclareRobustCommand{\ppp}[1]{\texttt{#1}}

\DeclareRobustCommand{\sqrtmeanvar}{\ppp{mean/var-sqrt}\xspace}
\DeclareRobustCommand{\meanvar}{\ppp{mean/var}\xspace}
\DeclareRobustCommand{\logmeanvar}{\ppp{mean/var-log}\xspace}
\DeclareRobustCommand{\nat}{\ppp{natural}\xspace}
\DeclareRobustCommand{\sqrtnat}{\ppp{natural-sqrt}\xspace}
\DeclareRobustCommand{\lognat}{\ppp{natural-log}\xspace}

\DeclareRobustCommand{\dataset}[1]{\textsc{#1}}

\usepackage{tikz, pgfplots, tikzexternal} 
\pgfplotsset{compat=newest}
\usepgfplotslibrary{groupplots}
\setstcolor{red}

 \newlength\figureheight
 \newlength\figurewidth
\begin{document}

\twocolumn[
\aistatstitle{Natural Gradients in Practice: Non-Conjugate  Variational Inference in Gaussian Process Models}
\aistatsauthor{Hugh Salimbeni \And Stefanos Eleftheriadis \And James Hensman}
\aistatsaddress{Imperial College London, PROWLER.io \And PROWLER.io \And PROWLER.io} 
]


\begin{abstract}
The natural gradient method has been used effectively in conjugate Gaussian process models, but the non-conjugate case has been largely unexplored. 
We examine how natural gradients can be used in non-conjugate stochastic settings, together with hyperparameter learning. 
We conclude that the natural gradient can significantly improve performance in terms of wall-clock time.
For ill-conditioned posteriors the benefit of the natural gradient method is especially pronounced, and we demonstrate a practical setting where ordinary gradients are unusable.
We show how natural gradients can be computed efficiently and automatically in any parameterization, using automatic differentiation. 
Our code is integrated into the GPflow package. 
\end{abstract}

\setlength\figureheight{8cm}
\setlength\figurewidth{.45\textwidth}

\begin{figure}[h]
	\small
\input{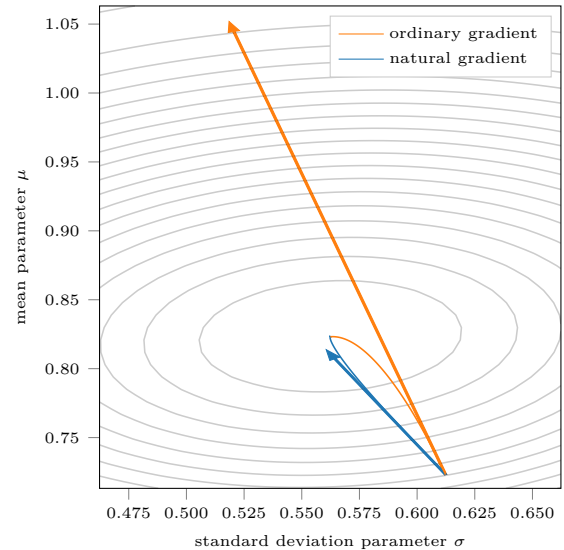}
  \caption{\small The natural gradient (blue arrow) is correctly scaled and points in a better direction than the ordinary gradient (orange arrow). The contours show the lower bound to a GP with a Bernoulli likelihood and variational posterior with a single Gaussian inducing point. The path followed by taking very small steps in the ordinary gradient (orange curve) ascends the contours. The path taking small steps in the  natural gradient (blue curve) is independent of parameterization, and does not follow the contours.}
\label{fig:gradients_with_contours}
\end{figure}

\setlength\figureheight{4cm}
\setlength\figurewidth{.33\textwidth}

\section{Introduction}
%
Minimizing the Kullback-Leibler ($\KL$) divergence between an unknown and a tractable parametric distribution is the central task of variational inference.
In the non-conjugate case, the prevalent approach is to optimize the objective using (stochastic) gradient descent or variants. 
Gradient descent based methods require careful tuning to work effectively and are prone to poor convergence when the Hessian at the solution is ill-conditioned~\citep{boyd2004convex}.
Ill-conditioning is a problem especially encountered in kernel methods~\citep{ma2017diving}. 
A further problem is that the \stepsize is not dimensionless but its units are in the square of the parameters.
The appropriate \stepsize is tightly coupled with the parameterization, and no best \stepsize can exist for all problems.

%
The ordinary gradient turns out to be an unnatural direction to follow for variational inference since we are optimizing a \emph{distribution}, rather than a set of parameters directly.
%
%
One way to define the gradient is the direction that achieves maximum change subject to a perturbation within a small euclidean ball.
To see why the euclidean distance is an unnatural metric for probability distributions, consider the two Gaussians $\mathcal N (0, 0.1)$ and $\mathcal N (0, 0.2)$, compared to $\mathcal N (0, 1000.1)$ and $\mathcal N (0, 1000.2)$.
The former pair are different and the latter similar, yet in euclidean distance they are equally far apart in the mean and variance.
Using the precision in place of the variance gives the opposite result, yet the distributions are unchanged.
There is a fundamental mismatch between the ordinary gradient and the objective function: the gradient is dependent on parameterization whereas the objective function is not.

Fortunately there is a way to solve the disparity: the natural gradient.
The natural gradient can be defined as the direction that achieves maximum change in $\KL$ divergence.
It is well known that paths following the natural gradient are invariant to reparameterization \cite[see e.g.,][]{martens2014new}, and that the natural gradient direction is the ordinary gradient rescaled by the inverse Fisher information matrix~\citep{amari1998natural}.
Fig.~\ref{fig:gradients_with_contours} shows a two parameter example comparing the natural gradient to the ordinary gradient.
In this case we see that the natural gradient points in a better direction than the ordinary gradient, and also has an appropriate scale.
%

To investigate whether the advantages suggested by Fig.~\ref{fig:gradients_with_contours} hold in practice, we consider several aspects in turn.
We begin by comparing the different gradients across several different parameterizations (\S \ref{section:linesearch}).
To achieve this we first demonstrate how natural gradients can be calculated efficiently and without any cumbersome derivations.
Through empirical investigation we show that the natural gradient is indeed a more effective direction to follow in all parameterizations, and also that there is an appropriate \stepsize to use after an initial phase.
Using these insights we propose a gradient descent algorithm for the common situation where the size of the dataset forces us to subsample the data, leading to stochastic gradients (\S~\ref{section:stochastic}).
We then extend the approach to hyperparameter optimization using a double loop algorithm that outperforms the state-of-the-art Adam optimizer in wall-clock time (\S \ref{section:hyperparameters}).
Finally, we demonstrate a situation where natural gradients are essential for successful optimization, due to ill-conditioning (\S \ref{section:ill_conditioning}).

Below we summarise our contributions:
\begin{itemize}
\item We compare natural gradients to other state-of-the-art techniques in non-conjugate problems, explicitly comparing the influence of different parameterizations and \stepsize on performance.   
\item We show how natural gradients in the exponential family can be computed efficiently and automatically in any parameterization, using automatic differentiation.
\item We show that natural gradients can be used in conjunction with hyperparameter learning in the stochastic setting.
\item We highlight a situation where the current approaches fail due to ill-conditioning, and show that natural gradients can solve this problem.
\item We provide an implementation of our methods within the GPflow~\citep{matthews2017gpflow} package.
\end{itemize}

\section{Background}
In this section we introduce the relevant background on the exponential family, variational inference, and optimisation approaches. We then define natural gradients and show how they take a simple form for the exponential family.
\subsection{Preliminaries}
We consider the problem of performing inference in a model of the form
\begin{align}
\label{eq:general_model}
p( \yv, \uv) = \left[ \prod_{n=1}^{N}p(y_n|\uv) \right]
p(\uv)\,,
\end{align}
where $y_n$ are observed and $\uv$ unobserved.
Both the prior and likelihood may additionally depend on hyperparameters, but we have omitted these from the notation to reduce clutter.
We assume that exact inference in~\eqref{eq:general_model} is intractable and make use of an approximate posterior $q(\uv ; \thetav)$ in the exponential family.
The exponential family is defined as
\begin{align}
\label{eq:exponential_family_def}
\log q(\uv;\thetav) =  \log h(\uv) + \thetav\T \Tv(\uv) - A(\thetav) \,, 
\end{align}
where $\Tv(\uv)$ is the sufficient statistics vector, $A(\thetav)$ is the log normalizing constant and $h(\cdot)$ is a base measure.
The parameterization used in~\eqref{eq:exponential_family_def} is known as the \emph{natural} parameterization and $\thetav$ are the natural parameters.
We can instead use an alternative smooth invertible parameterization $\xiv \equiv \xiv(\thetav)$.
We denote transformation between parameterizations through overloading notation, e.g., the inverse mapping back to the natural parameterization is $\thetav(\xiv)$, and we also abbreviate $q(\uv;\thetav(\xiv))$ as $q(\uv;\xiv)$. 
A parameterization of particular interest, known as the \emph{expectation} parameterization, is defined as $\etav \equiv \E_{q(\uv ; \thetav)}\Tv(\uv)$.
An important property of the exponential family is that the gradient of the log normalizer is equal to the expectation parameter: $\nabla_{\thetav} A(\thetav)=\etav\T$. 
This can be readily identified from differentiating \eqref{eq:exponential_family_def} with respect to $\thetav$ and taking expectations.\footnote{
Differentiating \eqref{eq:exponential_family_def}: $\nabla_{\theta} \log q(\uv ; \thetav) = \Tv(\uv)^\top - \nabla_{\thetav} A(\thetav)$.
Taking expectations: $\mathbb{E}_{q(\uv;\xiv)} \nabla_{\thetav}\log q(\uv;\thetav) =\etav^\top - \nabla_{\thetav} A(\thetav)$.
The score has expectation zero, so the result follows.
}
Differentiating~\eqref{eq:exponential_family_def} again, it follows that the Hessian of the log density is a Jacobian:
\begin{equation}
\label{eq:hessian_naturals}
\nabla^2_{\thetav} \log q(\uv;\thetav) = -\dd{\etav}{\thetav} \,.
\end{equation}
Variational inference proceeds by minimizing the $\KL$ divergence from $q(\uv;\xiv)$ to the intractable posterior $p(\uv|\yv)$, or equivalently maximizing the evidence lower bound (ELBO):
\begin{equation}
\label{eq:elbo}
\mathcal L(\xiv) = \mathbb{E}_{q(\uv;\xiv)} \sum_{n=1}^{N}\log p(y_n|\uv) - \KL[q(\uv;\xiv)||p(\uv)] \,.
\end{equation}
Our fundamental problem is to minimize $-\mathcal L(\xiv)$. All the approaches we consider find a sequence of parameters $\{\xiv_t\}_{t=0}^T$ using the iterative update
\begin{equation}
\label{eq:basic_update}
\xiv_{t+1} = \xiv_t - \gamma_t \Pv\inv_t \gv_t, \quad \gv_t = \nabla_{\xiv\T} \mathcal L \big\rvert_{\xiv=\xiv_t} \,,
\end{equation}
where $\gamma_t$ denotes the \emph{step size} and $\Pv\inv_t \gv_t$ the \emph{direction}.
%


\subsection{Optimization approaches}
%
\textbf{Gradient descent (GD)}. The simplest approach, known as gradient descent, is to set $\Pv$ to the identity matrix.
The step size can be fixed, decayed, or found by a line search on each iteration.
%
%

\textbf{Adam}. A more sophisticated approach is to use a diagonal matrix for $\Pv$, with diagonal elements given by $(\sqrt{v_i} + \epsilon) \inv m_i$, where $m_i$ and $v_i$ are the bias corrected exponential moving averages of $\left[\gv_t\right]_i$ and $\left(\left[\gv_t\right]_i\right)^2$.
This approach is called Adam~\citep{kingma2014adam}.

%
\textbf{\LBFGS}. One way of interpreting the update~\eqref{eq:basic_update} is to identify the term $\Pv\inv_t \gv_t$ as a minimizer of the local quadratic approximation $\mathcal L(\xiv_t + \deltav) \approx L(\xiv_t) + \gv\T\deltav + \tfrac12 \deltav\T\Pv\deltav$.
Under this interpretation, a natural choice for $\Pv$ is the Hessian so that the quadratic approximation coincides with the second order Taylor expansion.
This is known as Newton's method. Due to the large computational expense of calculating and inverting the Hessian we do not consider it further. Instead, we use a low rank approximation to the Hessian computed from finite differences.
Specifically we will compare to a common variant of this algorithm known as \LBFGS~\citep{byrd1995limited}.
This algorithm cannot be used in the stochastic setting as finite difference calculations are not robust to noise.

%
\textbf{Natural gradient descent (NGD)}. Another way of interpreting the update~\eqref{eq:basic_update} is to use the fact that the direction of steepest descent with respect to a norm $||\deltav||_{\mathbf A} = \deltav^T \mathbf A \deltav$ is given by $\mathbf A\inv \nabla_{\xiv\T} \mathcal L$.\footnote{This can be seen by minimizing $\tfrac{1}{\epsilon}\mathcal L(\xiv + \deltav)$ subject to the constraint that $||\deltav||_{\mathbf A} = \epsilon$ and letting $\epsilon \to 0$.}
Identifying $\Pv$ with $\mathbf{A}$, the update~\eqref{eq:basic_update} corresponds to the steepest descent with respect to the norm induced by the matrix $\Pv$.
Gradient descent (where $\Pv$ is the identity and the induced metric is Euclidean) can therefore be seen as moving in the direction that maximizes the change in objective with respect to the euclidean norm of the parameters.
The Euclidean norm is an unnatural way to compare two parameter vectors if the parameters correspond to \emph{distributions}, however.
If instead we consider the $\KL$ divergence between two distributions and take the small perturbation limit, we obtain $\KL[q(\uv ; \xiv), q(\uv ; \xiv + \deltav)] = \tfrac12 \deltav\T \left[ \E_{q(\uv ; \xiv)} \nabla_{\xiv}^2\log q(\uv ; \xiv) \right] \deltav + \mathcal O(||\deltav||^3)$. 
Therefore, in a sufficiently small neighbourhood the $\KL$ divergence induces a quadratic norm with curvature given by the expected Hessian of the log density.
This matrix is known as the Fisher information $\mathbf F_{\xiv}$,
\begin{align}
\label{eq:fisher_def}
\Fv_{\xiv} = -\E_{q(\uv ; \xiv)}\nabla^2_{\xiv}\log q(\uv ; \xiv) \,.
\end{align}
The direction of steepest descent with respect to this norm is called the natural gradient $\tilde{\nabla}_{\xiv}\mathcal L$, given by the gradient scaled by the inverse Fisher information: $\tilde{\nabla}_{\xiv} \mathcal{L}=\left( \nabla_{\xiv}\mathcal L \right)\mathbf F_{\xiv}^{-1}$ \citep{amari1998natural}.

For the exponential family the Fisher information takes a particularly simple form in the natural parameters. 
Using~\eqref{eq:hessian_naturals} we have that $\Fv_{\thetav} = \dd{\etav}{\thetav}$.
Using the chain rule, we see that the natural gradient in the natural parameters is given by $\tilde{\nabla}_{\thetav}\mathcal L=\dd{\mathcal L}{\etav}$. 
This expression was used by~\citet{Hensman2013} to compute natural gradients in the conjugate case. 

To find the natural gradients in some other paramterization we can use the chain rule to obtain
\begin{equation}
\label{eq:fisher_def_general}
\Fv_{\xiv} = \left(\dd{\thetav}{\xiv}\right)\T \dd{\etav}{\thetav} \dd{\thetav}{\xiv} \,.
\end{equation}
This expression was used directly in \citep{malago2015information} and \citep{sun2009efficient} in a certain parameterization of the Gaussian. The calculation is extremely cumbersome and requires a careful recursive implementation. In the next section we show how to compute the natural gradient efficiently and automatically.

\section{Efficient computation}
Since all the parameterizations are invertible (and the inverse of the Jacobian is the Jacobian of the inverse), we have 
\begin{align}
\label{eq:natural_gradient_xi}
\tilde{\nabla}_{\xiv} \mathcal L=& \dd{\mathcal L}{\xiv} \left( 
\left(\dd{\thetav}{\xiv} \right)\T
\dd{\etav}{\thetav} \dd{\thetav}{\xiv}\right)\inv\\
=& 
\dd{\mathcal L}{\xiv}  
\dd{\xiv}{\thetav} 
\dd{\thetav}{\etav} 
\left(\dd{\xiv}{\thetav}\right)\T \,.
\end{align}

Applying the chain rule and transposing, we obtain
\begin{align}
\label{eq:natural_gradient}
\tilde{\nabla}_{\xiv\T}\mathcal L = \dd{\xiv}{\thetav} \dd{\mathcal L}{\etav\T}  \,.
\end{align}
We recognise~\eqref{eq:natural_gradient} as a Jacobian-vector product, which is exactly what is computed in \emph{forward-mode} differentiation.
Forward-mode automatic differentiation libraries are perhaps less common than reverse-mode, but fortunately there is an elegant way to achieve forward-mode automatic differentiation using reverse-mode differentiation twice \citep{Townsend2017}. See the \supplementarymaterial for details on this trick.
Importantly, this indirect computation only costs negligibly more than the forward pass $\xiv(\thetav)$.
The extra computation comes from the parameter conversion between $\thetav$ and $\xiv$, which is $\mathcal O(M^3)$ for the (full rank) Gaussian for the six parameterizations we consider in the next section, where $M$ is the dimension of $\uv$.
Note that direct inversion of the Fisher information for the Gaussian would be cubic in the number of parameters, i.e., $\mathcal O((M + M^2)^3)$. 
In practice, we find this increases the computation relative to the ordinary gradient by a factor of about 1.5.  
We emphasize that this approach requires no more code than the parameter transformation, so new parameterizations can be easily investigated. 
%

\section{Specific application: Sparse Gaussian processes}
\label{section:parameterizations}
What we have described so far applies to any model and any exponential family variational posterior. 
We now present a specific example: a sparse Gaussian process (GP) model with a Gaussian variational posterior. For a comprehensive overview see~\citep{matthews2016sparse}.

The model takes the form of \eqref{eq:general_model}. Each $y_i$ is associated with a $D$-dimensional input $\mathbf x_i\in\R^D$. We place a GP prior on the unobserved variables $f(\mathbf x_i)$, 
\begin{align}
f(\cdot) \sim \mathcal{GP}(\mu, k) \,,
\end{align}
where $\mu$ and $k$ are mean and covariance functions. 
That is, any collection of function values $f(\mathbf x_1),\dots, f(\mathbf x_N)$ are jointly Gaussian with mean $\mu(\mathbf x_i)$ and covariance $k(\mathbf x_i, \mathbf x_j)$, for $i,j=1,\dots,N$. 
Inference in this model scales cubically in $N$, and is intractable when the likelihood is not Gaussian, so we proceed with variational inference. 
We choose a Gaussian process for the posterior with the special property that it matches the prior conditioned on a number of inducing points $\uv = [f(\zv_i)]_{i=1}^M$.
We use a directly parameterized Gaussian for $q(\uv)$. 
The posterior leads to the bound,
\begin{align}
\label{eq:lower_bound}
\mathcal L = \mathbb E_{q(\uv)} \sum_{i=1}^{N} \log \tilde p(y_i| \uv) - \KL[q(\mathbf u) || p(\mathbf u)] \,,
\end{align}
where $\log \tilde p(y_i | \uv)= \mathbb E_{q(f_i|\uv)} \log p(y_i| f_i) $. Since both expectations are over Gaussians they combine to a single expectation with mean and variance available in closed form. The univariate expectation of the likelihood can be found with Gauss-Hermite quadrature, or exactly in some cases. The bound \eqref{eq:lower_bound} can be evaluated stochastically by evaluating a random subset of terms in the sum and scaling the $\KL$ term appropriately.


\textbf{Parameterizations of the Gaussian}.
We now present the different parameterizations we will use of the Gaussian variational distribution $q(\uv)$.
The Gaussian is a member of the exponential family with the sufficient statistic vector given by $\Tv(\uv)=[\uv, \text{vec}(\uv\uv\T)]$, so that $\thetav \T \Tv(\uv) = \uv\T \thetav_1 +  \uv\T \boldsymbol\Theta_2\uv$, where $\thetav_1$ is the first $D$ elements of $\thetav$, and $\Thetav_2$ are remaining elements reshaped to a square matrix.
We refer to this as the \emph{unpacked form}.
A common parameterization of the Gaussian is in terms of the mean ($\mathbf m$) and variance ($\mathbf S$).
The unpacked natural parameters are given by $\mathbf S\inv \mathbf m, -\tfrac12 \mathbf S\inv$ and the expectation parameters by $\mathbf m, \mathbf S + \mathbf m \mathbf m\T$.
Converting between these parameterizations is straightforward and has complexity $\mathcal O (M^3)$.

We consider six parameterizations of the Gaussian. Perhaps the most commonly used in variational inference \citep[e.g.,][]{dai2015variational, challis2011concave} is the mean and square root of the covariance: $\mv, \mathbf{L}$, with $\mathbf{L}\mathbf{L}^T=\Sv$.
We refer to this as the \sqrtmeanvar parameterization.
Another way to constrain the covariance to be positive definite is to use the matrix log of the covariance \citep[e.g.,][]{glasmachers2010exponential} $\mv, \mathbf{L}$, with $\exp({\mathbf L})=\Sv$, where $\exp$ here is the matrix exponential (the \logmeanvar parameterization).
%
%
We use additionally the unconstrained mean and variance parameters $\mv, \Sv$ (the \meanvar parameterization) and the unconstrained natural parameters $\thetav_1, \Thetav_2$ (the \nat parameterization).
For completeness we also constrain the natural parameters via the square root and log transformations.
Since $\Thetav_2$ is negative definite we use $\thetav_1, \mathbf{L}$ with $ \mathbf{L} \mathbf{L}^T=-\Thetav_2$ (the \sqrtnat parameterization), and, finally,  $\thetav_1, \mathbf{L}$ with $\exp({\mathbf L})=-\Thetav_2$ (the \lognat parameterization).


%
%

\begin{figure*}[t]
\small
\centering
\input{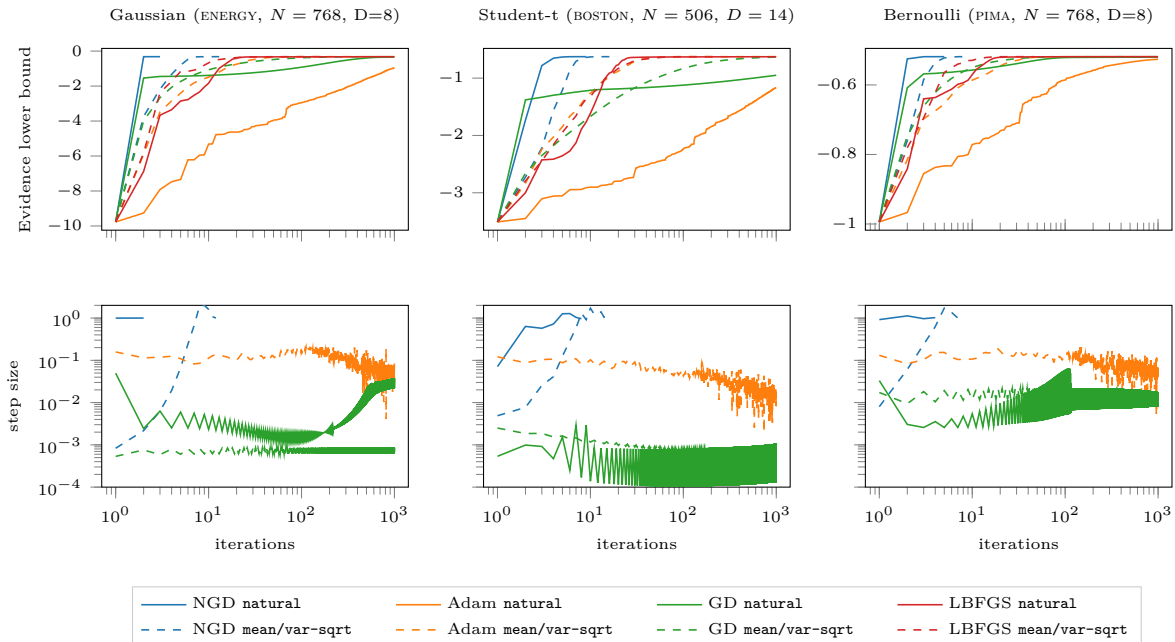}
  \caption{\small The natural gradient is a superior \emph{direction} in both parameterizations, and the best \stepsize increases during optimization to $\gamma\approx1$. Upper row: optimization methods, all with a \linesearch for the \stepsize. Lower row: the \stepsize{}s used at each iteration. Additional figures in the \supplementarymaterial confirm these claims with four further parameterizations and multiple splits.} \label{fig:deterministic_linesearch}
\end{figure*}

\section{Natural gradients in practice}
In this section we investigate NGD for large \stepsize{}s. We aim to provide evidence to answer the following:
\begin{enumerate}
\item Is the natural gradient a good \emph{direction}, irrespective of \stepsize?
\item Can we easily choose an effective \stepsize?
\item Are natural gradients useful when combined with hyperparameter optimization?
\end{enumerate}

We use a running example of three common datasets with different likelihoods: energy efficiency (\dataset{energy}, $N=784,D=8$) with a Gaussian likelihood, boston housing (\dataset{boston}, $N=506,D=14$) with a \studentT likelihood, and pima Indians diabetes (\dataset{pima}, $N=784,D=8$) with a Bernoulli likelihood.  
We use 100 inducing points initialized with k-means and the Matern ($\nu=\nicefrac{5}{2}$) kernel. Details of hyperparameters and data preprocessing are in the \supplementarymaterial. 

\subsection{Deterministic case}
\label{section:linesearch}
To investigate the quality of direction, we apply NGD, GD and Adam each with a \linesearch to find the $\gamma$ that achieves maximum value of the objective at each step.
We run an exhaustive search for $\gamma$ using the Brent~\citep{brent1971algorithm} method until convergence.
We compare also to \LBFGS which includes a \linesearch.
Plots are shown in the \supplementarymaterial for experiments using five splits of 90\% for each of the six parameterizations defined in~\S\ref{section:parameterizations}.
Fig.~\ref{fig:deterministic_linesearch} shows a representative split with the \sqrtmeanvar and \nat parameterizations.
For the case of the Gaussian likelihood we observe the optimal solution is found in a single step of $\gamma=1$, as shown in \citet{Hensman2013}.
For the other parameterizations the initial natural gradient \stepsize is a small value that is parameterization and likelihood dependent, but then increases to $\gamma=1$.
Once the \stepsize has increased to near $\gamma=1$ we observe extremely rapid convergence.

The natural gradient direction achieves faster convergence for all likelihoods and parameterizations (see \supplementarymaterial for the other 4 parameterizations).
%
%
%
For all different parameterizations and likelihoods we see that the best \stepsize for the natural gradient \emph{increases} to $\gamma=1$.
This is in contrast to the ordinary gradient where the \stepsize differs between likelihoods and generally needs to \emph{decrease} as optimization progresses.
For Adam the direction is \elementwise rescaled and a value close to 0.1 seems appropriate for the constrained parameterizations, but for the unconstrained parameterizations Adam cannot make good progress. 
%

In summary we have provided evidence that the natural gradient is indeed a better direction, and increasing the \stepsize to $\gamma=1$ is appropriate for fast convergence. 
We see also that the best combination of optimization method and parameterization is \nat for NGD and \sqrtmeanvar for GD and Adam. 
We will use these combinations for all subsequent experiments.

\subsection{Stochastic natural gradients}

\setlength\figureheight{5cm}
\setlength\figurewidth{.36\textwidth}
\begin{figure*}
\small
\centering
	\input{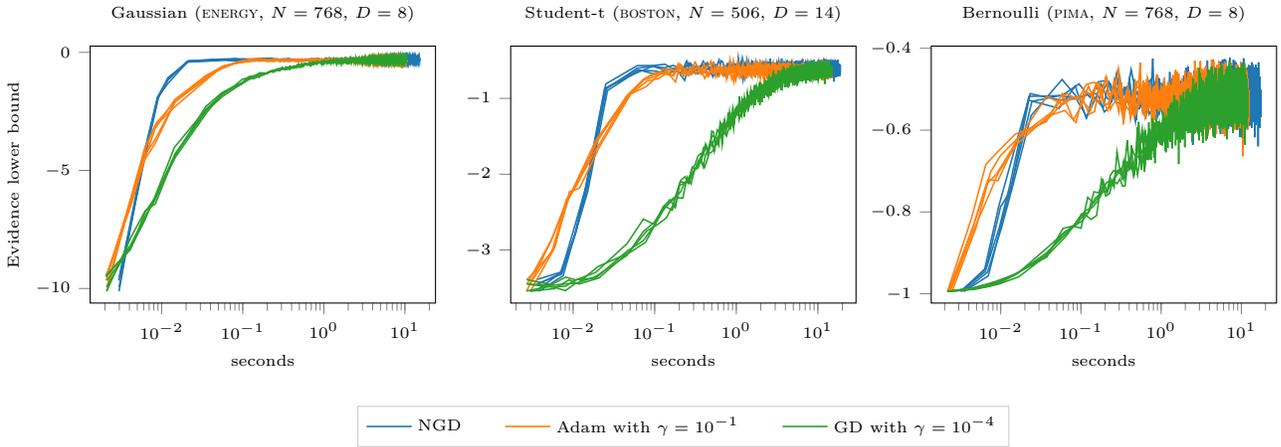}
  \caption{\small Stochastic optimization of the lower bound for fixed hyperparameters. The batch size is 256 and 5000 iterations are shown for five splits.}
  \label{fig:stochastic_fixed_step}
\end{figure*}

\begin{figure*}
\small
\centering
\input{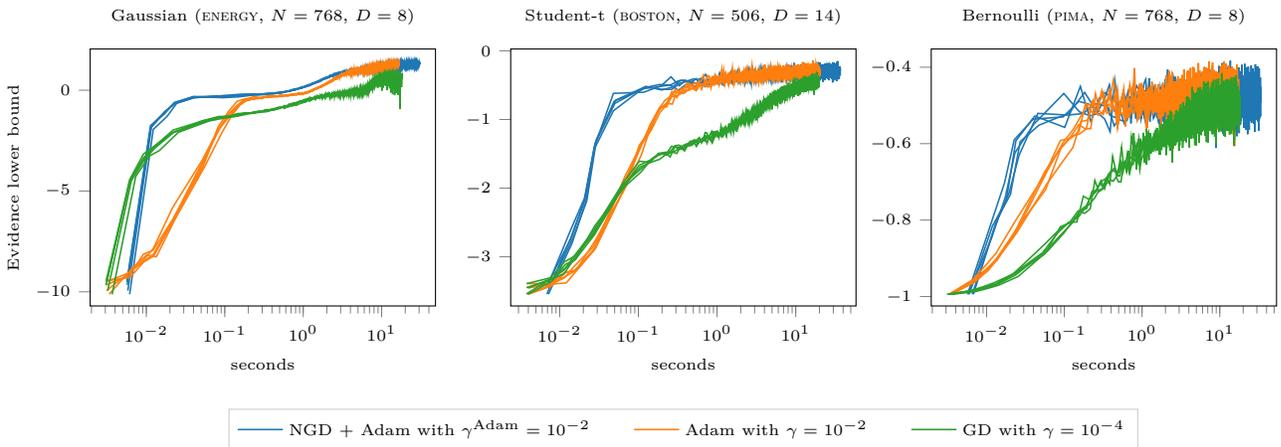}
  \caption{\small Joint optimization of the hyperparameters and the variational distribution. For natural gradients, a step of NGD on the variational parameters is alternated with a step of Adam on the hyperparameters. For Adam and GD the variational and hyperparameters are optimized together in a single objective. The batch size is 256 and 5000 iterations are shown for five splits.}
\label{fig:stochastic_hypers}
\end{figure*}

\begin{figure*}
\small
\input{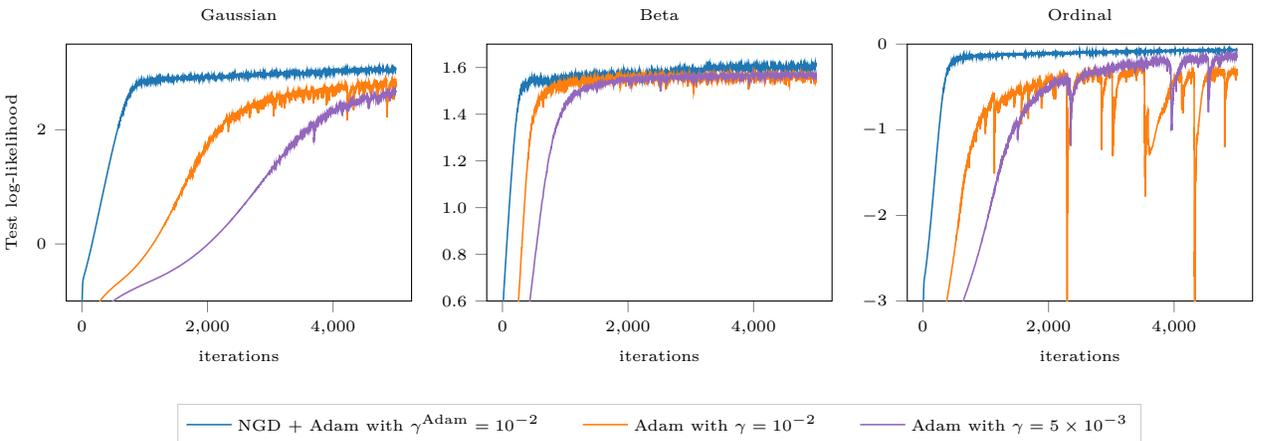}
	\caption{\small Optimization of the \dataset{naval}  dataset ($N=11$K, $D=16$), with three different likelihoods. The ill-conditioning of the variational distributions renders the optimization using ordinary gradients extremely difficult, even given a large number of iterations and different values for the Adam learning rate. The batch size is 256 and 5000 iterations are shown for a single split.}
\label{fig:naval}
\end{figure*}

\label{section:stochastic}
We next consider the stochastic case where a \linesearch is not possible. 
We introduce stochasticity by subsampling the data into minibatches of size 256.
To find a reasonable $\gamma$ for the Adam and GD methods we performed a search over $\{10^{-k}\}_{k=0}^{6}$.
We used the largest rate that remained stable. 

We now consider a strategy for $\gamma$. 
Our \linesearch experiments suggest that $\gamma$ should be gradually increased to some fixed value $\gamma\approx1$.
We therefore propose a simple schedule for NGD: (i) log-linearly increase $\gamma$ from $\gamma_{\text{initial}}$ to $\gamma_{\text{final}}$ over $K$ iterations; (ii) set $\gamma=\gamma_{\text{final}}$ for the remaining iterations. 
For the \dataset{energy, boston} and \dataset{pima} datasets we found that $\gamma_{\text{initial}}=10^{-4}$, $\gamma_{\text{final}}=10^{-1}$ and $K=5$ were suitable values. 
%

Fig.~\ref{fig:stochastic_fixed_step} shows the optimization of the \dataset{energy, boston} and \dataset{pima} datasets against wall-clock time with GD, Adam and NGD.
We observe that NGD improves on Adam and GD after about $3\times10^{-4}$ seconds (about 3 iterations). 
The advantages we see in the deterministic case appear to be realised in the stochastic setting. 

\subsection{Hyperparameters}
\label{section:hyperparameters}

An advantage of variational inference is that the ELBO can be optimized with respect to hyperparameters (we include also the inducing point inputs $\{\mathbf z_i\}_{i=1}^{M}$) as a proxy for the true marginal likelihood. 
Note that this is biased as the slack in the bound may depend on hyperparameter settings~\citep{turner2011two}. 
Nevertheless, it has been found to work well in practice, so a prevalent approach is to optimize the hyperparameters and variational parameters together in a single objective. 
We cannot use natural gradients directly for the hyperparameters as we do not have a probability distribution for them. 
Instead, we use an alternating scheme where we perform a step of Adam on the hyperparameters (with \stepsize $\gamma^{\text{Adam}}$), followed by a step of NGD on the variational parameters (with \stepsize $\gamma$)
We refer to this hybrid method as NGD+Adam and apply this approach to the same three datasets, using the same schedule for $\gamma$ as before. 
We compare to optimizing the variational distribution and hyperparmeters in a single objective using Adam and GD.

Fig.~\ref{fig:stochastic_hypers} shows the results of stochastic optimization of the variational distribution together with hyperparameters.
We see that NGD+Adam outperforms the other three methods in terms of wall-clock time.  
%


\subsection{When natural gradients are \emph{essential}}

\label{section:ill_conditioning}
In this section we present a practical situation where natural gradients are \emph{essential}. 
The previous experiments demonstrated settings where all approaches could find the same solution in a reasonable time.
This is not always the case, however, and we present a setting where the natural gradient approach can find a better solution than any method using ordinary gradient.  

In ill-conditioned settings 
 ordinary gradients suffer from instability \citep{sun2009efficient} and slow convergence.
As the natural gradient is invariant to parameterization, NGD is not adversely effected by issues of conditioning.
We consider the commonly used \dataset{naval} dataset, which has target values uniformly distributed in 51 increments between 0.95 and 1. 
We use this dataset with three different likelihoods: a Gaussian likelihood (rescaling the values to zero mean and unit variance), a single-parameter Beta likelihood\footnote{The usual $\alpha, \beta$ parameters are related by $\alpha=sm, \beta=s(1-m)$, with $m=\sigma(f)$ and $s>0$ a hyperparameter.} (rescaling to [0, 1]) and an ordinal likelihood (rescaling to $0, 1, \dots,  50$) \citep{chu2005gaussian}, with bins uniformly spaced between -2 and 2. 
For NGD+Adam use a schedule with $\gamma_{\text{initial}}=10^{-4}$, $\gamma_{\text{final}}=10^{-1}$ and $K=40$. 
We compare to the optimization of the lower bound with respect to hyperparameters and variational parameters using NGD$+$Adam and Adam. 

Fig.~\ref{fig:naval} shows the optimization progress in terms of test log-likelihood after a large number of iterations.
Note that ordinary gradient with Adam cannot achieve the optimal value, even after many iterations and with different \stepsize{s}.

\subsection{Further Results}
\setlength\figureheight{5cm}
\setlength\figurewidth{.27\textwidth}
\begin{figure*}[ht]
\small
\centering
\input{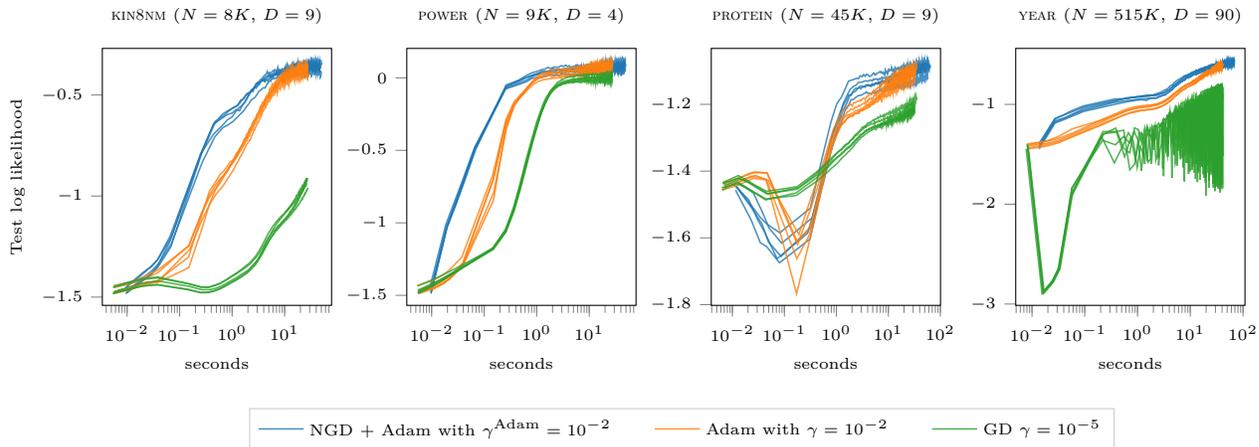}
  \caption{\small Optimization of hyperparameters and variational distributions for larger UCI datasets with a \studentT likelihood.}
\label{fig:uci}
\end{figure*}

\setlength\figurewidth{.45\textwidth}
\begin{figure}
\small
\centering
\input{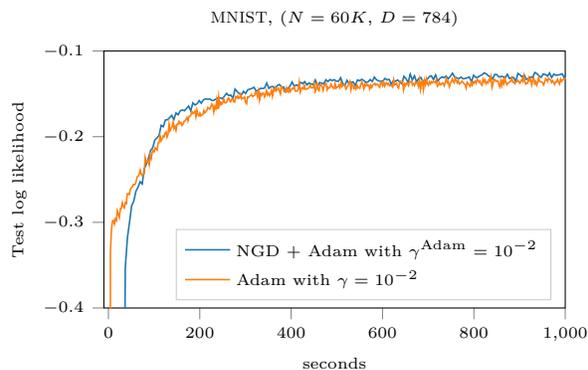}
  \caption{\small Optimization of the hyperparameters and variational distribution for the \dataset{mnist} data, with a Robust-max multiclass likelihood. We see that after the first few initial iterations NGD+Adam outperforms Adam alone.}
  \label{fig:mnist}
\end{figure}

In this section we apply NGD$+$Adam to 4 larger datsets from UCI corpus, using a \studentT likelihood, and also the \dataset{mnist} for multiclass classification. 
In both settings we find that natural gradients either find the optimal solution more quickly, or enable a solution to be found that cannot be obtained using ordinary gradients alone.

Fig.~\ref{fig:uci} shows the optimization in the UCI datasets with \studentT likelihood, using a minibatch size of 256 and with the same schedule for $\gamma$ as in the \dataset{naval} experiment. 
For the \dataset{kin8nm}, \dataset{power} and \dataset{year} datasets we observe significant improvement over Adam.

Fig.~\ref{fig:mnist} shows the result of \dataset{mnist} multiclass classification using the standard train/test split and a batch size of 1024. 
The schedule for $\gamma$ increases log-linearly from $10^{-6}$ to $0.02$ over 2000 iterations. 
We see that the natural gradient approach outperforms Adam in terms of test loglikelihood.

\section{Related work}
The first use of natural gradients for variational inference goes back to~\citet{sato2001online}, where it was shown that for an exponential family conditionally conjugate model (i.e., a model where classical fixed point variational updates can be derived in closed form), the NGD corresponds exactly to the fixed point variational update if $\gamma=1$. 
This observation leads to an online version of the fixed point algorithm. 
This idea was made more explicit in~\citep{Hoffman2013}, where it was termed stochastic variational inference and applied to a range of problems. 
The first example of natural gradients used in the non-conjugate case with Gaussian variational distribution can be found in~\citep{honkela2010approximate}.
In this work, natural gradients are used for the variational mean.  
The inverse Fisher information for the mean has a particularly simple form (it is the precision), but the expression for the covariance is much more complicated and cumbersome to derive directly. 

Natural evolution strategies (NES)~\citep{sun2009efficient} is closely related to variational inference. 
In NES a fitness function is optimized in expectation under a Gaussian. 
This converges to a zero entropy solution, so ordinary gradients cannot feasibly be used due to the problem of ill-conditioning.
Natural gradients are therefore essential for a practical algorithm.
In~\citep{sun2009efficient} the Fisher information for the \sqrtmeanvar is calculated and inverted directly, which is inefficient. 
A similar result for the \meanvar parameterization was presented in~\citep{malago2015information}.

Recently, there have been several works employing natural gradients to approximations of non-conjugate components of a model. 
In the context of GPs, \citet{khan2015kullback,khan2016faster}  used a linearization of the non-conjugate terms and achieved impressive results. 
\citet{johnson2016composing} use an auxiliary model to learn the approximate natural parameters with neural network likelihood, and then perform analytic updates on the conjugate approximation. 
\citet{knowles2011non} use model-specific bounds to take approximate natural gradient steps in a variational message passing setting. 

\section{Discussion and conclusion}
In all cases that we have investigated, we found that natural gradients accelerate convergence relative to methods using the ordinary gradient. 
In some cases the contrast is so severe that the ordinary gradient can require an unfeasibly large number of iterations to achieve the same results as the natural gradient.
In practice, natural gradients are essential for finding a good solution in these situations. 
The drawbacks of the approach are that a schedule for $\gamma$ must be specified.
The success of the method relies on $\gamma$ increasing to a reasonably large value ($\approx0.1$) sufficiently quickly ($<1000$ iterations). 
If $\gamma$ needed to be kept small for much longer, then the advantage of the natural gradient method might be lost. 
Using a probabilistic \linesearch \citep{mahsereci2015probabilistic} for NGD is a promising area for future research.

We have shown that natural gradients are useful for variational inference in non-conjugate sparse Gaussian process models.
Natural gradients are particularly advantageous in problems where the ordinary gradient is crippled by the parameterization-dependent ill-conditioning.
Such situations exist in practice.
We have shown that natural gradients can be computed efficiently and with minimal effort using modern automatic differentiation techniques, and can be combined with modern optimizers such as Adam for hyperparameter learning.
We compared six likelihoods and nine benchmark datasets, and found the natural gradient provided improvement in all cases.

\subsubsection*{Acknowledgements}
We have greatly appreciated valuable discussions with Mark van der Wilk in the preparation of this work.
\bibliographystyle{plainnat}
\bibliography{references}
\newpage
\onecolumn

\section*{Supplementary Material}

\setlength\figureheight{4cm}
\setlength\figurewidth{.33\textwidth}
\subsection{Further \linesearch figures}

\begin{figure*}[!ht]
\small
\centering
\input{figs/line_search_nat.tex}
  \caption{\small Line search in the \nat parameterization across 5 splits.}
\end{figure*}

\begin{figure*}[!ht]
\small
\centering
\input{figs/line_search_sqrtnat.tex}
  \caption{\small Line search in the \sqrtnat parameterization across 5 splits.}
\end{figure*}

\begin{figure*}[!ht]
\small
\centering
\input{figs/line_search_lognat.tex}
  \caption{\small Line search in the \lognat parameterization across 5 splits.}
\end{figure*}

\begin{figure*}[!ht]
\small
\centering
\input{figs/line_search_meanvar.tex}
  \caption{\small Line search in the \meanvar parameterization across 5 splits.}
\end{figure*}

\begin{figure*}[!ht]
\small
\centering
\input{figs/line_search_sqrtmeanvar.tex}
  \caption{\small Line search in the \sqrtmeanvar parameterization across 5 splits.}
\end{figure*}

\begin{figure*}[!ht]
\small
\centering
\input{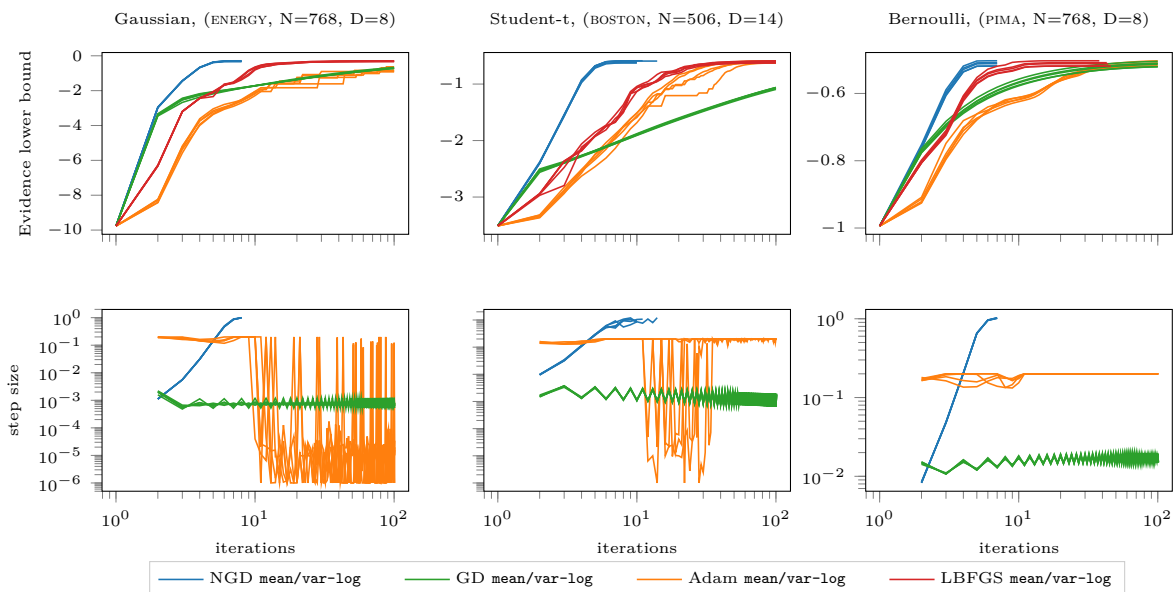}
  \caption{\small Line search in the \logmeanvar parameterization across 5 splits.}
\end{figure*}

\clearpage
\subsection{Experimental details}
\textbf{Implementation}.
All experiments were run on a single desktop machine with a GTX 1070 GPU. The code was written in GPflow~\citep{matthews2017gpflow}, a GP library built on tensorflow. 

\textbf{Kernel}.
For all experiments we used the Matern $\frac52$ kernel, with the (single) lengthscale initialized to the square root of the data dimension. The kernel variance was initalized to 2, except for \dataset{mnist}, where we initialized to 10.

\textbf{Inducing points}.
We used 100 inducing points, initialized with k-means. The variational parameters were intialized to mean zero and identity covariance.

\textbf{Jitter}.
We used a small jitter level of $10^{-10}$ for all experiments. 

\textbf{Data normalization}.
For all datasets apart from \dataset{mnist} we scaled the inputs to have zero mean and unit variance in the training data. We applied the same scaling to the test data. For \dataset{mnist} we used the standard scaling to the unit interval.

For the Gaussian and \studentT likelihoods we scaled the outputs to have zero mean and unit standard deviation in the training data. The beta and ordinal likelihood are described in the main text.

\subsection{The forward-mode trick}
We describe how to obtain a forward-mode derivative using a reverse-mode library. The trick is due to \citet{Townsend2017} and this explanation closely follows \url{https://j-towns.github.io/2017/06/12/A-new-trick.html}. 

Reverse-mode differentiation is the successive application of the vector-Jacobian product (vjp) operation.  
The vjp operation left multiplies a vector $\mathbf u$ with the Jacobian of $\mathbf f$ with respect to its input $\mathbf x$:
\begin{align*}
\text{vjp}(\mathbf f, \mathbf x, \mathbf u) = \mathbf u\T \dd{\mathbf f}{\mathbf x} = \sum_i u_i \dd{f_i}{\mathbf x} \,.
\end{align*}

The vjp operation can be used to implement the gradient of a function $L(\mathbf f(\mathbf g(\mathbf x)))$ by using the chain rule $\dd{L}{\mathbf x}=\dd{L}{\mathbf f}\dd{\mathbf f}{\mathbf g}\dd{\mathbf g}{\mathbf x}$ and  successively applying the vjp operation from left to right, i.e., 
\begin{align*}
\mathbf u & = \text{vjp}(L , \mathbf f, 1)\,, \\
\mathbf u & \leftarrow \text{vjp}(\mathbf f, \mathbf g, \mathbf u)\,, \\
\mathbf u & \leftarrow \text{vjp}(\mathbf g, \mathbf x, \mathbf u) \,.
\end{align*}
After these operations $\mathbf u = \dd{L}{\mathbf x}$. Automatic reverse-mode differentiation libraries implement vjp for all basic operations they support. Compositions of basic operations can be computed as above. Note that the values of $\mathbf f$ and $\mathbf g$ need to be computed first, which requires a forward pass through the function.

Forward-mode differentiation makes use of a Jacobian-vector product operation (jvp), defined as 
\begin{align*}
\text{jvp}(\mathbf f, \mathbf x, \mathbf u) =\dd{\mathbf f}{\mathbf x}  \mathbf u = \sum_i \dd{\mathbf f}{x_i} u_i \,.
\end{align*}

Using the jvp operation, the chain rule can be implemented by successive application of jvp, working from right to left, i.e.,
\begin{align*}
\mathbf u & = \text{jvp}(\mathbf g , \mathbf x, \mathbf 1)\,, \\
\mathbf u & \leftarrow \text{jvp}(\mathbf f, \mathbf g, \mathbf u)\,, \\
\mathbf u & \leftarrow \text{jvp}(L, \mathbf f, \mathbf u) \,,
\end{align*}
where $\mathbf 1$ is a vector of ones with the same shape as $\mathbf x$. 

To implement natural gradients in any parameterization we require the jvp operation, but common libraries such as Tensorflow implement only vjp (i.e. reverse mode). The trick to achieve jvp from vjp is to introduce a dummy variable $\mathbf v$ and define $\mathbf g(\mathbf v) = \text{vjp}(\mathbf f, \mathbf x, \mathbf v)$. We then use vjp again to find the gradient of $\mathbf g$ with respect to $\mathbf v$, passing in the vector $\mathbf u$ to be pushed forward: $\text{vjp}(\mathbf g, \mathbf v, \mathbf u)$. Since $\mathbf g$ is linear in $\mathbf v$ , we have 
\begin{align*}
\text{vjp}(\mathbf g, \mathbf v, \mathbf u) = \mathbf u^{\top}\dd{}{\mathbf v}\left(\mathbf v^{\top} \dd{\mathbf f}{\mathbf x} \right) =  \mathbf u^{\top} \left(\dd{\mathbf f}{\mathbf x}\right)^{\top}\,.
\end{align*}
This is exactly the transpose of $\text{jvp}(\mathbf f, \mathbf x, \mathbf u)$. Therefore, any reverse-mode differentiation library can be used to compute forward-mode derivatives.

\end{document}